\documentclass{article} 
\usepackage{iclr2027_conference,times}

\usepackage{amsmath,amsfonts,bm}

\def\eqref#1{equation~\ref{#1}}

\def\1{\bm{1}}

\DeclareMathAlphabet{\mathsfit}{\encodingdefault}{\sfdefault}{m}{sl}
\SetMathAlphabet{\mathsfit}{bold}{\encodingdefault}{\sfdefault}{bx}{n}

\DeclareMathOperator{\sign}{sign}

\usepackage{amssymb}
\usepackage{url}

\usepackage{epigraph}
\usepackage{algorithm, algorithmic}
\usepackage{booktabs}
\usepackage{graphicx}
\usepackage{hyperref}
\usepackage{multirow}

\title{ShamAN-Q: Shampoo Augmented NanoQuant for Sub-1-bit LLM Weights}

\author{Jonathan Mei, Sang Hyub Kim, Oliver Knitter, Chi Chen \& Martin Roetteler \\
Applications R\&D\\
IonQ\\
College Park, MD 20740, USA \\
\texttt{\{jmei,sang,oliver.knitter,chi.chen,martin.roetteler\}@ionq.co}
}

\iclrfinalcopy 
\begin{document}

\providecommand{\method}{ShamAN-Q}
\providecommand{\Diag}{\operatorname{Diag}}
\providecommand{\tr}{\operatorname{tr}}
\providecommand{\sign}{\operatorname{sign}}

\maketitle
\lhead{Preprint}

\begin{abstract}
We introduce \method{}, a sub-1-bit post-training quantization method that
extends NanoQuant by replacing each its diagonal reconstruction geometry with a
tractable dense curvature metric, using a general
paradigm popularized by the Shampoo optimizer.  For each linear weight,
\method{} fits a Kronecker product to the empirical Fisher
information matrix of a small calibration set by Kullback--Leibler
minimization, forming a Mahalanobis
reconstruction loss from the result.  The continuous ADMM updates from NanoQuant become solutions to
Sylvester equations, while its discrete projection and deployment format remain unchanged.  Because the curvature is local to
a given set of weights, \method{} re-measures the input curvature statistic for each layer
immediately before layer factorization, periodically refreshing all statistics
on the partially quantized model. \method{} also redistributes the uniform
rank from NanoQuant across layers at the same total number of bits.  On Qwen3-Base,
\method{} lowers WikiText-2 perplexity at $\approx$1\,bpw from 27.56 to 22.96
(0.6B), 19.21 to 16.72 (1.7B), and 14.29 to 13.80 (4B) while matching or improving zero-shot accuracy on the Eleuther LM Evaluation Harness. On 0.6B, \method{} at $\approx$0.8\,bpw matches the published perplexity of NanoQuant at $\approx$1.0\,bpw.
\end{abstract}


\section{Introduction}
\label{sec:intro}

\begin{figure}[!b]
\centering
\includegraphics[width=\linewidth]{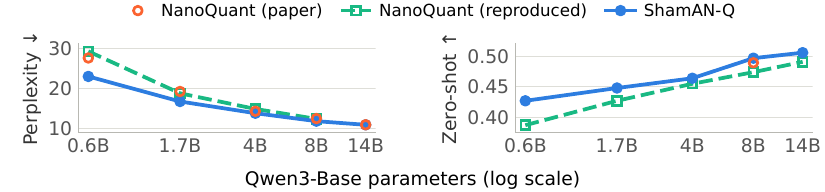}
\caption{WikiText-2 perplexity (PPL) (left, lower is better) and zero-shot mean on reasoning tasks in Eleuther LM Evaluation Harness
(right, higher is better) vs.\ Qwen3-Base size at a 1.0\,bpw budget. \method{} matches or improves on NanoQuant in both PPL and zero-shot mean at every size.
All points are from Table~\ref{tab:main}; only reported results are plotted.}
\label{fig:ppl-vs-size}
\end{figure}

The energy consumption of language models is an omnipresent, growing
concern, affecting all scales from those of frontier data centers down
to small edge devices.  Awareness of this cost grows alongside demand for
data sovereignty, independence from model providers, and fast local
intelligence. Beyond the cost of the computation itself, an often overlooked source of language model
energy cost is the repeated movement of data.
For causal language model inference, processing a prompt in parallel during
prefill is often compute-bound. A long prompt produces a large effective
batch size $B$, and assuming $n\times n$ matrices, the resulting matrix--matrix multiplication moves
$O(n^2)$ data for $O(Bn^2)$ computation.  Decoding,
by contrast, is often bandwidth-bound, since each generated token requires
transferring the entire active set of parameters from memory to the compute
cores, before counting any host--device or device--device transfers required
by models that do not fit on a single accelerator.  In the extreme, batch
size $B=1$ reduces to matrix--vector multiplication with $O(n^2)$ data
movement for only $O(n^2)$ computation.

Shrinking the stored weights therefore directly attacks the dominant cost
of decoding, and large language models are increasingly deployed under memory
and bandwidth budgets that their weights alone exceed. Post-training
quantization (PTQ) to 4 bits per weight (bpw) is now well established \citep{frantar2022gptq}, and even 2 bpw is maturing
\citep{lin2024awq}, but the \emph{sub-1-bit} regime requires a
structural change, since a scalar codebook cannot spend less than one bit on every
weight. NanoQuant \citep{nanoquant2025} instead represents each decoder
linear layer by two low-rank binary factors and two short FP16 scale vectors
(Eq.~\ref{eq:deployed-weight}). By doing so, NanoQuant changes the dominant storage from one
bit per matrix entry to $r(m+n)$ bits per layer at factor rank $r\gg 16$.  At
this compression level, nearly every weight is perturbed, and as a result accuracy depends
not only on the size of the reconstruction error but also on which input and
output directions absorb it.

NanoQuant handles unequal channel scales through diagonal
preconditioning, after which its optimization process minimizes a Frobenius reconstruction loss. This diagonal preconditioning acts as a proxy for second-order information specific to each channel, but it does not capture second-order relationships between channels. \method{} extends NanoQuant by replacing the channel-wise
scaling with a dense metric that captures these channel correlations. In this regard, \method{} is to NanoQuant what Shampoo \citep{gupta2018shampoo} is to
diagonal AdaGrad \citep{duchi2011adagrad}. One-sided Hessian-aware PTQ methods such as GPTQ \citep{frantar2022gptq}
model input sensitivity while treating all output directions alike.
The Fisher Information matrix of a weight matrix captures both, but its $mn\times mn$ size is prohibitive
for an $m\times n$ weight.  Approximating it with an $n\times n$ input
statistic and an $m\times m$ output statistic makes this information tractable
\citep{martens2015kfac} (\S\ref{sec:bg-kron}).

\method{} estimates dense curvature statistics from a small calibration set
while limiting the influence of massive-activation tokens. As quantization changes the model, these statistics are regularized and
refreshed; they simultaneously serve to guide
low-rank reconstruction and rank allocation at NanoQuant's total bit
budget, thus preserving its binary representation and deployment format.
At this budget, \method{} lowers WikiText-2 perplexity on Qwen3-Base from
27.56 to 22.96 at 0.6B, 19.21 to 16.72 at 1.7B, and 14.29 to 13.80 at 4B while matching or improving zero-shot accuracy on the Eleuther LM Eval Harness; at 0.8\,bpw, \method{} at 0.6B matches NanoQuant's published 1.0\,bpw perplexity
(\S\ref{sec:experiments}).  We now describe the primary contributions of this paper:
\begin{itemize}
\item A dense two-sided reconstruction metric extension of NanoQuant for sub-1-bit PTQ: KL-Shampoo
  fit, and a fresh input statistic with periodic full
  refresh (\S\ref{sec:curvature}, \S\ref{sec:block-recon}), resulting in a modified ADMM whose continuous updates are solutions to Sylvester
  equations. Other components, such as the Sign-Value Independent Decomposition (SVID) projection \citep{pouransari2020least, xu2020low}, dual updates, scale extraction, and packed format structures remain unchanged (\S\ref{sec:block-recon}).
\item A redistribution of NanoQuant's uniform rank budget at bit parity from
  a combination of depth, layer type, and a measurement of reconstruction loss for each layer
  (\S\ref{sec:rank-alloc}).
\item Consistent decreases in perplexity paired with increases in zero-shot reasoning performance over NanoQuant on Qwen3-Base from 0.6B to 14B at identical $\lesssim 1$-bit budgets, attributed to each component
  (\S\ref{sec:experiments}).
\end{itemize}

\section{Related Work}
\label{sec:related}
\textbf{Binary and Sub-1-bit Quantization of LLMs.}
Weight-only PTQ is already an established paradigm at 4 and 2 bpw
\citep{frantar2022gptq,lin2024awq}. Pushing further, binary methods either train with
quantization-aware objectives, as in the cases of BitNet \citep{wang2023bitnet} and
OneBit \citep{xu2024onebit}, or partially binarize and keep salient weights
at higher precision, as in the case of PB-LLM \citep{shang2023pbllm};
\citet{binaryllm2025}, in contrast, revisits 1-bit optimization from pre-trained weights. Generally, in-place binarization with full-precision scales is bounded below by one
bit per weight, and grouping metadata pushes the effective rate higher.
NanoQuant \citep{nanoquant2025} breaks this bound in the PTQ setting by
factorizing each weight into two low-rank binary matrices and two scale
vectors, initialized by a latent-binary ADMM under diagonal preconditioning
and refined block by block (\S\ref{sec:bg-nanoquant}).

\textbf{Hessian-aware PTQ and Kronecker Curvature Statistics.}
GPTQ \citep{frantar2022gptq} demonstrates a form of Hessian-aware PTQ by weighting the reconstruction
loss with the input second moment, a one-sided Hessian surrogate that models
input sensitivity while treating all output directions alike. The second order optimizers K-FAC
\citep{martens2015kfac} and Shampoo \mbox{\citep{gupta2018shampoo}} approximate the
layer Fisher by a Kronecker product of input and output statistics, and
BaKron \citep{bakron2025} applies similar Kronecker-structured Hessians to weight
quantization. Two estimators fit the Kronecker product to the empirical
Fisher directly, the nearest Kronecker product (NKP) in Frobenius norm
\citep{mei2023kradagrad,morwani2025} and the matrix-normal
maximum-likelihood fit of KL-Shampoo \citep{klshampoo2025,proklshampoo2026}.
\S\ref{sec:bg-kron} presents both, and in \method{} we adopt the KL fit
because of how it weights massive-activation tokens \citep{sun2024massive}.

\textbf{ADMM for Structured Compression.}
The alternating direction method of multipliers \citep{boyd2011admm} suits
objectives that pair a continuous reconstruction loss with a hard discrete
constraint.  NanoQuant introduces LB-ADMM, which decouples the low-rank
reconstruction from the binary structure by using SVID-projected consensus
variables. \method{} retains this effective decoupling and its Euclidean consensus
penalties, so that only the continuous subproblems change. These subproblems change from right-sided
normal equations (Eq.~\ref{eq:nq-normal-eq}) to two-sided Sylvester
equations.

\section{Background}
\label{sec:background}
\subsection{NanoQuant}
\label{sec:bg-nanoquant}

\textbf{Representation and bit cost.}
NanoQuant \citep{nanoquant2025} deploys each linear layer, with weight
$\mathbf W\in\mathbb R^{m\times n}$ and factor rank $r$, in the two-scale
form
\begin{equation}
\widehat{\mathbf W}
=\Diag(\mathbf s_1)\,\mathbf U_{\pm1}\,
 \mathbf V_{\pm1}^{\top}\,\Diag(\mathbf s_2),
\qquad
\mathbf U_{\pm1}\in\{\pm1\}^{m\times r},\quad
\mathbf V_{\pm1}\in\{\pm1\}^{n\times r},
\label{eq:deployed-weight}
\end{equation}
using FP16 scale vectors of lengths $m$ and $n$.  Since the signs require
$r(m+n)$ bits, and the scales $16(m+n)$ bits, a rank-$r$ layer costs
\begin{equation}
\operatorname{bits}(r)=(r+16)(m+n),
\qquad
\operatorname{BPW}
=(r+16)(m+n)/mn.
\label{eq:bpw}
\end{equation}

\textbf{Pipeline.}
Phase~1, \emph{global calibration}, runs a calibration set through the
full-precision teacher $\mathcal M$ and collects, per linear layer and
token $t$, the layer input $\mathbf x_t\in\mathbb R^{n}$ and output
gradient $\boldsymbol\delta_t\in\mathbb R^{m}$.  Phase~2, \emph{block
reconstruction}, processes decoder blocks in order. Within a block, Step~1
(\textsc{TuneFP}) tunes the remaining full-precision weights to reproduce
the teacher's block output from the already-quantized prefix's activations,
and is repeated between layer factorizations. Step~2 initializes each
linear layer's binary factors and scales by the LB-ADMM below. Step~3
(\textsc{TuneLatentSTE}) jointly tunes the block's continuous latents
$\mathcal U,\mathcal V$ and scales $\mathbf s_1,\mathbf s_2$ under a
weighted MSE block loss with the straight-through estimator
\citep{bengio2013ste}, after which $\mathbf U_{\pm1}=\sign(\mathcal U)$ and
$\mathbf V_{\pm1}=\sign(\mathcal V)$ are packed.  Phase~3
(\textsc{TuneScalesKD}), tuning only the scale vectors of every layer, minimizes the forward KL divergence between the next-token distributions of the teacher and quantization models.

\textbf{Diagonal preconditioning (Step 2-1).}
NanoQuant performs one measurement of the per-input-feature
activation second moments and the per-output-feature output gradient second moments for the full-precision model, after robust token clipping, and shrinks them
toward their mean,
\begin{equation}
\mathbf d_{\mathrm{in}}^{2}
=\frac1N\sum_{t}\mathbf x_t\odot\mathbf x_t,
\qquad
\mathbf d_{\mathrm{out}}^{2}
=\frac1N\sum_{t}\boldsymbol\delta_t\odot\boldsymbol\delta_t,
\qquad
\mathbf d^{2}\leftarrow(1-\gamma)\,\mathbf d^{2}
+\gamma\,\operatorname{mean}(\mathbf d^{2})\,\mathbf 1,
\label{eq:nq-precond}
\end{equation}
with $\gamma\in[0,1]$. These shrunken moments then produce
$\widetilde{\mathbf D}_{\mathrm{in}}=\Diag(\mathbf d_{\mathrm{in}})$,
$\widetilde{\mathbf D}_{\mathrm{out}}=\Diag(\mathbf d_{\mathrm{out}})$, the
transformed target
$\widetilde{\mathbf W}
=\widetilde{\mathbf D}_{\mathrm{out}}\mathbf W\widetilde{\mathbf D}_{\mathrm{in}}$,
and the loss
$\lVert\widetilde{\mathbf E}\rVert_F^{2}
=\tr\big(\mathbf E^{\top}\widetilde{\mathbf D}_{\mathrm{out}}^{2}
\mathbf E\widetilde{\mathbf D}_{\mathrm{in}}^{2}\big)$ where $\mathbf E=\mathbf W-\widehat{\mathbf W}$ and
$\widetilde{\mathbf E}
=\widetilde{\mathbf D}_{\mathrm{out}}\mathbf E\widetilde{\mathbf D}_{\mathrm{in}}$.
This loss comprises the two-sided metric of \S\ref{sec:bg-kron} with diagonal statistics.

\textbf{LB-ADMM (Step 2-2).}
The continuous optimization variables are $\mathbf U\in\mathbb R^{m\times r}$ and
$\mathbf V\in\mathbb R^{n\times r}$, and they correspond with consensus variables
$\mathbf Z_U$ and $\mathbf Z_V$, which are restricted to the SVID family
$\mathcal C_{d,r}=\{\Diag(\mathbf a)\mathbf S\Diag(\mathbf b):
\mathbf a\in\mathbb R_+^d,\ \mathbf b\in\mathbb R_+^r,\
\mathbf S\in\{\pm1\}^{d\times r}\}$.  NanoQuant then solves
\begin{equation}
\min_{\substack{\mathbf U,\mathbf V,\mathbf Z_U,\mathbf Z_V\\
\mathbf U=\mathbf Z_U,\ \mathbf V=\mathbf Z_V}}\;
\frac12\lVert\widetilde{\mathbf W}-\mathbf U\mathbf V^\top\rVert_F^{2}
+\frac{\lambda}{2}
 \left(\lVert\mathbf U\rVert_F^{2}+\lVert\mathbf V\rVert_F^{2}\right)
+\mathcal I_{\mathcal C_{m,r}}(\mathbf Z_U)
+\mathcal I_{\mathcal C_{n,r}}(\mathbf Z_V)
\label{eq:nq-admm-objective}
\end{equation}
by a scaled-dual ADMM, with ridge $\lambda$ and a penalty $\rho_k$ that grows
linearly over $K$ iterations.  Taking $\alpha_k=\rho_k+\lambda$, with the consensus
target $\mathbf C_U^k=\mathbf Z_U^k-\boldsymbol\Lambda_U^k$ being formed using the
scaled dual $\boldsymbol\Lambda_U^k$, the $\mathbf U$ update is the
right-sided $r\times r$ normal equation
\begin{equation}
\mathbf U^{k+1}
\big((\mathbf V^k)^{\top}\mathbf V^k+\alpha_k\mathbf I_r\big)
=\widetilde{\mathbf W}\mathbf V^k+\rho_k\mathbf C_U^k,
\label{eq:nq-normal-eq}
\end{equation}
while the $\mathbf V$ update is analogous. As shown for $\mathbf Z_U$, the consensus variables are
projected according to
$\mathbf Z_U^{k+1}=\textsc{SVID}(\mathbf U^{k+1}+\boldsymbol\Lambda_U^k)$,
where SVID takes the sign of its argument and the three largest singular values of
its absolute value, and the scaled duals are updated as
$\boldsymbol\Lambda_U^{k+1}
=\boldsymbol\Lambda_U^k+\mathbf U^{k+1}-\mathbf Z_U^{k+1}$
(Appendix~\ref{app:algorithms}).

\textbf{Scale extraction (Step 2-3).}
The final proxies $\mathbf U^K+\boldsymbol\Lambda_U^K$ and
$\mathbf V^K+\boldsymbol\Lambda_V^K$ are magnitude-balanced. Their signs
are become $\mathbf U_{\pm1},\mathbf V_{\pm1}$, their row-wise mean
absolute values give the scales, and the inverse of the diagonal maps return the
scales to the original coordinates (Appendix~\ref{app:algorithms}).

\subsection{Kronecker Curvature Statistics: Shampoo and KL-Shampoo}
\label{sec:bg-kron}

\textbf{Empirical Fisher and the two-sided metric.}
For calibration token $t$, the weight gradient is the rank-one matrix
$\boldsymbol\delta_t\mathbf x_t^{\top}$, and the empirical layer Fisher information, in
the original coordinates of $\operatorname{vec}(\mathbf W)$, is given by
$\widehat{\mathbf H}_{\mathrm{cal}}
=\sum_t\operatorname{vec}(\boldsymbol\delta_t\mathbf x_t^{\top})
\operatorname{vec}(\boldsymbol\delta_t\mathbf x_t^{\top})^{\top}
\in\mathbb R^{mn\times mn}$.
A Kronecker approximation can estimate this matrix as
$\widehat{\mathbf H}_{\mathrm{cal}}\approx\mathbf A\otimes\mathbf G$ using an
input-side statistic $\mathbf A\in\mathbb R^{n\times n}$ and an output-side
statistic $\mathbf G\in\mathbb R^{m\times m}$
\citep{martens2015kfac,gupta2018shampoo}.  We call these
\emph{curvature statistics} and reserve \emph{factors} for the low-rank weight
matrices $\mathbf U$ and $\mathbf V$.  Under
column-major vectorization, $\mathbf A$ and $\mathbf G$ act respectively on input and output coordinates, and the metric induced on an error
$\mathbf E$ is
\begin{equation}
\left\lVert\mathbf E\right\rVert_{\mathbf G,\mathbf A}^{2}
\mathrel{:=}
\tr\!\left(\mathbf E^{\top}\mathbf G\mathbf E\mathbf A\right)
=\operatorname{vec}(\mathbf E)^{\top}
 (\mathbf A\otimes\mathbf G)\operatorname{vec}(\mathbf E).
\label{eq:kron-metric}
\end{equation}
All computations use the two statistics directly, so the full $mn\times mn$ Kronecker
product never needs to be formed.  NanoQuant uses as its loss the diagonal case
$\mathbf A=\widetilde{\mathbf D}_{\mathrm{in}}^{2}$,
$\mathbf G=\widetilde{\mathbf D}_{\mathrm{out}}^{2}$, which is equivalent to
$\mathbf A=\mathbf I_n$, $\mathbf G=\mathbf I_m$ in the transformed
coordinates.

\textbf{Shampoo.}
Shampoo \citep{gupta2018shampoo} preconditions a matrix gradient
$\nabla_t=\boldsymbol\delta_t\mathbf x_t^{\top}$ with two accumulated
statistics,
\begin{equation}
\mathbf G_{\mathrm{Sh}}
=\sum_t\nabla_t\nabla_t^{\top}
=\sum_t\lVert\mathbf x_t\rVert^{2}\,
  \boldsymbol\delta_t\boldsymbol\delta_t^{\top},
\qquad
\mathbf A_{\mathrm{Sh}}
=\sum_t\nabla_t^{\top}\nabla_t
=\sum_t\lVert\boldsymbol\delta_t\rVert^{2}\,
  \mathbf x_t\mathbf x_t^{\top},
\label{eq:shampoo}
\end{equation}
and updates according to
$\mathbf W\leftarrow\mathbf W
-\eta\,\mathbf G_{\mathrm{Sh}}^{-1/4}\,\nabla\,\mathbf A_{\mathrm{Sh}}^{-1/4}$,
which corresponds in vectorized form with the preconditioner
$(\mathbf A_{\mathrm{Sh}}^{1/2}\otimes\mathbf G_{\mathrm{Sh}}^{1/2})^{-1/2}$.
\citet{morwani2025} show that the square of Shampoo's approximation is one
power-iteration step toward the optimal Kronecker approximation of
$\widehat{\mathbf H}_{\mathrm{cal}}$, so the exponent $1/4$ per statistic
is the inverse square root of an approximate curvature.

\textbf{Estimators as token-weighted second moments.}
Kronecker fits of $\widehat{\mathbf H}_{\mathrm{cal}}$ all share the form
$\mathbf G\propto\sum_t w^{x}_t\,\boldsymbol\delta_t\boldsymbol\delta_t^{\top}$,
$\mathbf A\propto\sum_t w^{\delta}_t\,\mathbf x_t\mathbf x_t^{\top}$, differing only in their token weights.  The nearest-Kronecker-product (NKP) fit
\citep{mei2023kradagrad,morwani2025} minimizes
$\lVert\widehat{\mathbf H}_{\mathrm{cal}}-\mathbf A\otimes\mathbf G\rVert_F$
over positive semidefinite statistics, with fixed-point weights
$w^{x}_t=\mathbf x_t^{\top}\mathbf A\,\mathbf x_t$,
$w^{\delta}_t=\boldsymbol\delta_t^{\top}\mathbf G\,\boldsymbol\delta_t$;
Shampoo's statistics coem from its first alternation from
$\mathbf A=\mathbf I_n$, $\mathbf G=\mathbf I_m$.  The \textsc{KL-Shampoo}
fit \citep{klshampoo2025,proklshampoo2026} is the matrix-normal
maximum-likelihood estimate
\begin{equation}
(\mathbf A,\mathbf G)
\in\arg\min_{\mathbf A,\mathbf G\succ\mathbf 0}
D_{\mathrm{KL}}\!\left(
  \mathcal N(\mathbf 0,\widehat{\mathbf H}_{\mathrm{cal}})
  \,\middle\|\,
  \mathcal N(\mathbf 0,\mathbf A\otimes\mathbf G)
\right),
\qquad
\tr(\mathbf A)=n,
\label{eq:kl-fit}
\end{equation}
where the trace constraint removes the scale ambiguity
$(c\mathbf A)\otimes(c^{-1}\mathbf G)=\mathbf A\otimes\mathbf G$.
Stationarity then gives the coupled fixed point
\begin{equation}
\mathbf G
=\frac1n\sum_{t}\big(\mathbf x_t^{\top}\mathbf A^{-1}\mathbf x_t\big)\,
  \boldsymbol\delta_t\boldsymbol\delta_t^{\top},
\qquad
\mathbf A
=\frac1m\sum_{t}\big(\boldsymbol\delta_t^{\top}\mathbf G^{-1}\boldsymbol\delta_t\big)\,
  \mathbf x_t\mathbf x_t^{\top},
\label{eq:kl-fixed-point}
\end{equation}
so the token weights are the \emph{inverse} quadratic forms.  The
KL-Shampoo optimizer tracks these statistics by an exponential moving average
and updates
$\mathbf W\leftarrow\mathbf W-\eta\,\mathbf G^{-1/2}\,\nabla\,\mathbf A^{-1/2}$,
with exponent $1/2$ per statistic because $\mathbf A\otimes\mathbf G$ now
approximates $\widehat{\mathbf H}_{\mathrm{cal}}$ directly.

\section{\method{}}
\label{sec:method}
\begin{figure}[!t]
\centering
\includegraphics[width=\linewidth]{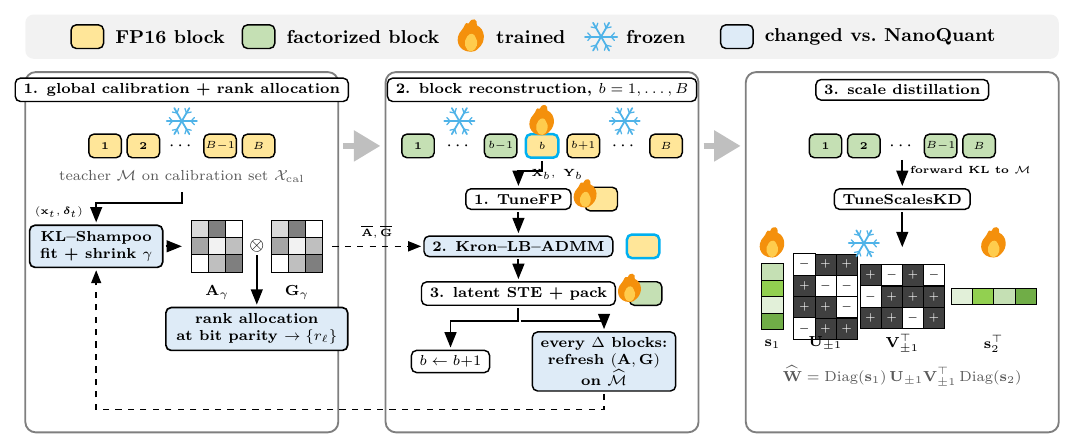}
\caption{\method{} end to end.  Phase~1 runs the calibration set through
the frozen teacher, fits and shrinks the curvature statistics
$(\mathbf A_\gamma,\mathbf G_\gamma)$ of every linear layer, and allocates
ranks at bit parity.  Phase~2 processes decoder blocks in order. The
factorized prefix and the untouched full-precision suffix are frozen while
block $b$ is tuned (\textsc{TuneFP}), factorized layer by layer by
Kron--LB--ADMM under the transported statistics
$(\overline{\mathbf A},\overline{\mathbf G})$ (Figure~\ref{fig:dense-admm}),
and refined by latent STE. The statistics of the
unfactorized layers are refreshed on the partially quantized model after every $\Delta$ blocks.
Phase~3 distills only the scale vectors against the teacher, with the sign
matrices frozen. Flames mark trained parameters, snowflakes frozen ones,
and blue blocks denote differences between \method{} and NanoQuant.}
\label{fig:pipeline}
\end{figure}

\begin{figure}[!ht]
\centering
\includegraphics[width=\linewidth]{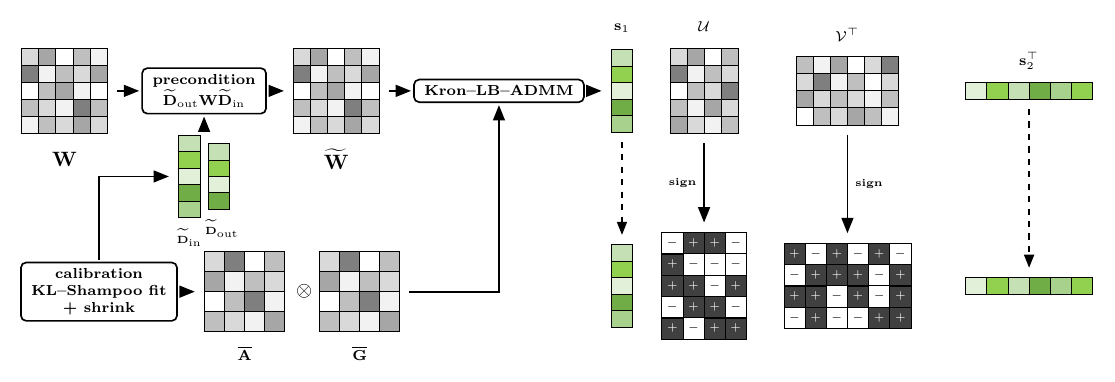}
\caption{\method{} pipeline for one layer.  Calibration fits and shrinks the
curvature statistics, yielding the transported metric
$(\overline{\mathbf A},\overline{\mathbf G})$ and diagonal maps
$(\widetilde{\mathbf D}_{\mathrm{in}},\widetilde{\mathbf D}_{\mathrm{out}})$.
Preconditioning maps $\mathbf W$ to $\widetilde{\mathbf W}$. Kron--LB--ADMM
returns the continuous weight factors $(\mathcal U,\mathcal V)$ and scale
vectors $(\mathbf s_1,\mathbf s_2)$.  The transported statistics
alter only the continuous $\mathbf U/\mathbf V$ optimization component; Euclidean consensus
leaves SVID, dual updates, scale extraction, and sign extraction unchanged
from NanoQuant.}
\label{fig:dense-admm}
\end{figure}

\method{} keeps the same representation as NanoQuant and similarly adopts a three-phase pipeline
(\S\ref{sec:bg-nanoquant}), but changes the geometry used to fit the low-rank factors at four interfaces. Curvature is estimated during the initial calibration
(\S\ref{sec:curvature}), used during rank allocation (\S\ref{sec:rank-alloc}), applied to the
continuous factor updates, and refreshed with respect to
the partially quantized model (\S\ref{sec:block-recon}).  All components related to SVID, scale
extraction, latent tuning, packing, and scale distillation are unchanged.
Figure~\ref{fig:pipeline} shows the three phases and marks their changed
stages, while Figure~\ref{fig:dense-admm} details the curvature-aware ADMM
fitting of one layer, which is the primary extension over NanoQuant. The complete pipeline is given in Appendix~\ref{app:algorithms}.

\subsection{Dense Kronecker Curvature from Global Calibration}
\label{sec:curvature}

\paragraph{Curvature estimation.}
Where NanoQuant keeps only the diagonal of
$\widehat{\mathbf H}_{\mathrm{cal}}$ (Eq.~\ref{eq:nq-precond}), \method{}
keeps its full Kronecker structure; $(\mathbf A,\mathbf G)$ are fitted for each
linear layer by three alternating passes of the KL-Shampoo fixed point
(Eq.~\ref{eq:kl-fixed-point}) over the calibration set, with Cholesky
inverses. The KL fit is preferred to NKP for its token weighting
(\S\ref{sec:bg-kron}). The NKP weights
$\mathbf x_t^{\top}\mathbf A\,\mathbf x_t$ are quartic in activation
magnitude, so the few tokens corresponding with massive activations \citep{sun2024massive}
dominate the leading eigenvalues. The KL weights
$\mathbf x_t^{\top}\mathbf A^{-1}\mathbf x_t$, in contrast, are bounded leverage scores,
and the resulting statistics have condition numbers of order 40 instead of
several hundred.  The estimator consumes only the pairs
$(\mathbf x_t,\boldsymbol\delta_t)$ and never materializes
$\widehat{\mathbf H}_{\mathrm{cal}}$.

\paragraph{Scale-matched shrinkage.}
The fitted statistics are regularized before any coordinate change,
\begin{equation}
\mathbf A_\gamma=(1-\gamma)\mathbf A+\gamma\,\tfrac{\tr(\mathbf A)}{n}\,\mathbf I_n,
\qquad
\mathbf G_\gamma=(1-\gamma)\mathbf G+\gamma\,\tfrac{\tr(\mathbf G)}{m}\,\mathbf I_m,
\qquad \gamma\in[0,1].
\label{eq:kron-shrink}
\end{equation}
Each identity has the same mean eigenvalue as the statistic it regularizes, so
shrinkage changes anisotropy without arbitrarily changing its scale:
there is no additional damping term.  The diagonal of the shrunken statistic is
NanoQuant's shrunken $\mathbf d^{2}$ of Eq.~\ref{eq:nq-precond}, and
NanoQuant's robust token clipping is kept when the statistics are
accumulated.  The fit and shrinkage of this section form the calibration
stage of Figure~\ref{fig:dense-admm}. They run once at calibration and at
each periodic refresh, while the diagonal maps, transport, and tempering of
\S\ref{sec:block-recon} run per layer.

\subsection{Non-uniform Rank Allocation at Bit Parity}
\label{sec:rank-alloc}

NanoQuant gives every layer the same bits per weight: setting
$\operatorname{BPW}=\beta$ in Eq.~\ref{eq:bpw} yields
\begin{equation}
r^{\star}_\ell
=\beta\,m_\ell n_\ell / (m_\ell+n_\ell)-16,
\qquad\beta=1,
\label{eq:uniform-rank}
\end{equation}
giving $r^{\mathrm{uni}}_\ell$ when floored to a multiple of 32.  \method{} keeps
the total
$N_{\mathrm{bits}}=\sum_\ell\operatorname{bits}_\ell(r^{\mathrm{uni}}_\ell)$
(bit parity) and redistributes it across layers in steps of 32 with rank
ceiling $\min(m_\ell,n_\ell)$.  Given $b_\ell\in\{1,\dots,B\}$, the block of
layer $\ell$ and its type $t(\ell)$ (q, k, v, o, gate, up, down), both
allocators use the depth prior
$\omega_\ell=\exp\big(\kappa\big(\tfrac{b_\ell-1}{B-1}-\tfrac12\big)\big)\,w_{t(\ell)}$,
with $\kappa=0.6$, so the last block gets more bits than the first by a factor of $e^{0.6}\approx1.8$.  The \emph{hand table} sets budgets
$\beta_\ell\propto\beta\,\omega_\ell$---where the type weights $w_t$ are contained in a range from 0.85 (q)
to 1.15 (down), taken from the block-loss jump each layer type caused when
binarized in the 4B logs---and rounds them to ranks meeting exactly
$N_{\mathrm{bits}}$.  The \emph{measured sensitivity} allocator
instead probes each layer with a short dense-curvature LB-ADMM
(\S\ref{sec:block-recon}) at three ranks, scores the result by the
Gauss--Newton error
$J_\ell(r)=\lVert\mathbf W_\ell-\widehat{\mathbf W}_\ell(r)\rVert^{2}
_{\mathbf G_{\gamma,\ell},\mathbf A_{\gamma,\ell}}$ using the layer's own
metric, fits a power law, multiplies its level by the depth prior with all
$w_t=1$, and hands out 32-rank steps greedily by predicted loss decrease per
bit.  The probes rank the seven layer types within a block well, but
under-value late blocks, which the prior supplies.  Appendix~\ref{app:rank}
gives the type weights, rounding, and probe protocol.

\subsection{Block Reconstruction Pipeline}
\label{sec:block-recon}
\label{sec:locality}

Blocks are processed in order, and within each block every linear layer is
factorized according to NanoQuant's per-layer schedule of
\S\ref{sec:bg-nanoquant}.  Step~1 (\textsc{TuneFP}) and Step~3
(\textsc{TuneLatentSTE}, sign extraction, and packing) are unchanged;
\method{} changes only the preconditioning and the continuous solves inside
Step~2.

\textbf{Periodic curvature refresh.}
The Fisher information is local to the full-precision weights, and the model drifts from
that point as blocks are binarized.  \method{} corrects for this at two time
scales: a per-layer fresh input statistic inside Step~2-1, and a periodic
refresh at block boundaries. For every $\Delta$ blocks, the statistics of all \emph{not-yet-factorized} layers are re-estimated with one warm-started pass of
Eq.~\ref{eq:kl-fixed-point} through the current partially quantized model
$\widehat{\mathcal M}$, so that later blocks see the statistics they will
actually receive, and are shrunk with $\gamma$ as before.

\textbf{Step 2-1: Dense-Curvature Preconditioning.}
\label{sec:precondition}
NanoQuant measures $\widetilde{\mathbf D}_{\mathrm{in}}$ once, on the
full-precision model (Eq.~\ref{eq:nq-precond}).  Under the block-wise schedule, by the time layer
$\ell$ is factorized, its inputs come from an already quantized prefix and
from full-precision layers that \textsc{TuneFP} has just modified.
Immediately before ADMM for each layer, we therefore re-measure its input
second moment on exactly those inputs,
$\mathbf A_{\mathrm{fresh}}
=\frac1N\sum_{t=1}^{N}\mathbf x_t\mathbf x_t^{\top}$, tkaen
over the $N$ calibration tokens with no gradient weighting.  Shrunken by
Eq.~\ref{eq:kron-shrink}, it then takes the place of $\mathbf A_\gamma$ for that
layer, while the output-side statistic $\mathbf G_\gamma$ is that of either the
calibration time or most recent refresh.

The diagonal preconditioners are the square roots of the diagonals of the
shrunken statistics, floored at $\epsilon>0$ to keep both maps invertible
(Eq.~\ref{eq:diag-coordinates}), while $\widetilde{\mathbf W}$ and
$\widetilde{\mathbf E}$ are taken as in \S\ref{sec:bg-nanoquant}.  Rather than
refitting in these coordinates, we transport the same shrunken statistics by
congruence, which preserves the Mahalanobis error exactly:
\begin{equation}
\overline{\mathbf A}
=\widetilde{\mathbf D}_{\mathrm{in}}^{-1}
 \mathbf A_\gamma
 \widetilde{\mathbf D}_{\mathrm{in}}^{-1},
\qquad
\overline{\mathbf G}
=\widetilde{\mathbf D}_{\mathrm{out}}^{-1}
 \mathbf G_\gamma
 \widetilde{\mathbf D}_{\mathrm{out}}^{-1},
\qquad
\left\lVert\mathbf E\right\rVert_{\mathbf G_\gamma,\mathbf A_\gamma}^{2}
=
\left\lVert\widetilde{\mathbf E}
\right\rVert_{\overline{\mathbf G},\overline{\mathbf A}}^{2},
\label{eq:metric-transport}
\end{equation}
which is obtained by substituting
$\mathbf E=\widetilde{\mathbf D}_{\mathrm{out}}^{-1}
\widetilde{\mathbf E}\widetilde{\mathbf D}_{\mathrm{in}}^{-1}$
into Eq.~\ref{eq:kron-metric} and cycling the trace. Notwithstanding the floor, the transported statistics are unit-diagonal
correlation matrices, so the dense metric combines NanoQuant's diagonal
preconditioning with $(\overline{\mathbf G},\overline{\mathbf A})$. Thus
NanoQuant is recovered exactly by $\overline{\mathbf A}=\mathbf I_n$ and
$\overline{\mathbf G}=\mathbf I_m$.

The one-bit reconstruction error is large
($\lVert\mathbf E\rVert\approx0.4$--$0.5\,\lVert\mathbf W\rVert$), well
outside the neighborhood in which the Fisher is an accurate curvature, so
before the ADMM solver is applied, each transported statistic is replaced by its
trace-preserving power (\emph{spectral tempering}).  Writing
$\overline{\mathbf A}=\mathbf Q\Diag(\mathbf c)\mathbf Q^{\top}$, we then have
\begin{equation}
\overline{\mathbf A}
\;\longleftarrow\;
\frac{\sum_i c_i}{\sum_i c_i^{\,p}}\;
\mathbf Q\Diag(\mathbf c^{\,p})\mathbf Q^{\top},
\qquad p=\tfrac12,
\label{eq:tempering}
\end{equation}
and likewise for $\overline{\mathbf G}$.  This halves the log-condition
number, and as a result, the data term trusts the dominant curvature directions less. The
square root is the regret-optimal metric under an uncertain curvature.
Tempering is free given the cached eigendecomposition. It leaves the identity
unchanged, and beats both $p=1$ and Shampoo's $p=\tfrac14$
(\S\ref{sec:exp-ablation}).  All uses of
$\overline{\mathbf A},\overline{\mathbf G}$ below refer to the tempered
statistics; their eigensystems and
$\mathbf M\mathrel{:=}\overline{\mathbf G}\,\widetilde{\mathbf W}\,\overline{\mathbf A}$
are cached per layer (Algorithm~\ref{alg:prepare-metric} in
Appendix~\ref{app:algorithms}).

\textbf{Step 2-2: Dense-Curvature LB-ADMM.}
\label{sec:kron-admm}
Relative to NanoQuant's objective (Eq.~\ref{eq:nq-admm-objective}),
\method{} replaces the Frobenius reconstruction term by the transported
metric and keeps everything else.  In the transformed coordinates it solves
\begin{align}
\min_{\mathbf U,\mathbf V,\mathbf Z_U,\mathbf Z_V}\quad
&\frac12
\left\lVert
  \widetilde{\mathbf W}-\mathbf U\mathbf V^\top
\right\rVert_{\overline{\mathbf G},\overline{\mathbf A}}^2
+\frac{\lambda}{2}
 \left(\left\lVert\mathbf U\right\rVert_F^2
      +\left\lVert\mathbf V\right\rVert_F^2\right)
+\mathcal I_{\mathcal C_{m,r}}(\mathbf Z_U)
+\mathcal I_{\mathcal C_{n,r}}(\mathbf Z_V)
\nonumber\\
&\text{subject to}\quad
\mathbf U=\mathbf Z_U,\quad\mathbf V=\mathbf Z_V.
\label{eq:kron-admm-objective}
\end{align}
with $\mathbf Z_U$ and $\mathbf Z_V$ in the SVID family of
\S\ref{sec:bg-nanoquant}.  The reconstruction term is Mahalanobis, but the
two augmented consensus penalties are Euclidean.  This keeps SVID, dual
updates, the $\rho_k$ schedule, and residuals identical to NanoQuant, so
only the continuous solves change.  Figure~\ref{fig:dense-admm} follows one
layer from calibration through preconditioning, the continuous
Kron--LB--ADMM solve, and scale and sign extraction;
Algorithm~\ref{alg:kron-admm} in Appendix~\ref{app:algorithms} lists the
solver in full.

Take $\alpha_k=\rho_k+\lambda$ and let
$\mathbf C_U^k=\mathbf Z_U^k-\boldsymbol\Lambda_U^k$ be the consensus target
formed with the scaled dual $\boldsymbol\Lambda_U^k$.  Holding
$\mathbf V^k$ fixed, stationarity of Eq.~\ref{eq:kron-admm-objective} gives
\begin{equation}
\overline{\mathbf G}\,\mathbf U^{k+1}\mathbf R_V^k
+\alpha_k\mathbf U^{k+1}
=\mathbf M\mathbf V^k+\rho_k\mathbf C_U^k,
\qquad
\mathbf R_V^k=(\mathbf V^k)^\top\overline{\mathbf A}\mathbf V^k,
\label{eq:u-sylvester}
\end{equation}
and the $\mathbf V$ update is obtained analogously, by exchanging the roles of
$(\overline{\mathbf G},\mathbf U)$ and $(\overline{\mathbf A},\mathbf V)$ and using $\mathbf M^\top$ in place of $\mathbf M$.  These are
Mahalanobis Sylvester equations, rather than NanoQuant's right-sided
$r\times r$ normal equations (Eq.~\ref{eq:nq-normal-eq}). Because the
augmented penalties are Euclidean, the left curvature statistic does not
cancel.  Both have the form
$\mathbf C\mathbf X\mathbf R+\alpha\mathbf X=\mathbf B$, with $\mathbf C$ a
tempered, transported curvature statistic and
$\mathbf R\in\mathbb R^{r\times r}$ positive semidefinite.  With the
curvature eigensystem $\mathbf C=\mathbf Q\Diag(\mathbf c)\mathbf Q^\top$
cached once per layer and
$\mathbf R=\mathbf S\Diag(\boldsymbol\nu)\mathbf S^\top$ recomputed for each
update, the solution in the joint eigenbasis is
\begin{equation}
\mathbf X=\mathbf Q\mathbf Y\mathbf S^\top,
\qquad
\mathbf Y_{i,j}
=\left[\mathbf Q^\top\mathbf B\mathbf S\right]_{i,j}/(c_i\nu_j+\alpha).
\label{eq:sylvester-solve}
\end{equation}
Because $\alpha=\rho+\lambda>0$, every denominator is positive even for
semidefinite statistics. We observe that taking $\overline{\mathbf G}=\mathbf I_m$ and
$\overline{\mathbf A}=\mathbf I_n$ recovers Eq.~\ref{eq:nq-normal-eq}. One
eigendecomposition per statistic per layer comprises the added cost
(Appendix~\ref{app:complexity}), and the SVID projection and its inner
iteration count are independent of
$\overline{\mathbf A},\overline{\mathbf G}$
(Appendix~\ref{app:algorithms}).

\textbf{Step 2-3: Unchanged Scale Extraction.}
\label{sec:scales}
The magnitude balancing and scale extraction that NanoQuant uses are applied verbatim
to the final proxies in the transformed coordinates, followed by its
un-preconditioning through the inverse diagonal maps
(Appendix~\ref{app:algorithms}). The resulting sign matrices initialize the
latents $\mathcal U,\mathcal V$ of Step~3, and the block loss,
packing, and \textsc{TuneScalesKD} all come from NanoQuant as well
(\S\ref{sec:bg-nanoquant}).

\section{Experiments}
\label{sec:experiments}
\subsection{Experimental Setup}
\label{sec:exp-setup}

\textbf{Implementation and Environment.}
\method{} is implemented on top of the released NanoQuant code and follows its
protocol wherever \S\ref{sec:method} does not change it. The 0.6B to 4B results use a single NVIDIA
A100-80GB GPU with fp64 eigendecompositions. The 8B and 14B results use fp32 eigendecompositions on two to four GPUs, which was verified to be lossless at 1.7B.
Hyperparameters and the engineering measures that keep the dense recipe
within budget are given in Appendix~\ref{app:impl}.

\textbf{Models and Datasets.}
We quantize Qwen3-Base at 0.6B, 1.7B, and 4B, 8B, and 14B parameters \citep{qwen3-2025},
calibrating on 128 WikiText-2 \citep{merity2016pointer} sequences of 2048
tokens (seed 0). At 8B and 14B the curvature statistics and their refreshes use 512 sequences, while block reconstruction and distillation still use 128. We report both WikiText-2 perplexity at context 2048 and the mean zero-shot accuracy from the Eleuther LM Evaluation Harness \citep{gao2023lmeval}.
All numbers are single runs.

\textbf{Baselines.}
The baseline is NanoQuant at the same total number of bits, presented both as
published \citep{nanoquant2025} and as our own paper-faithful reproduction.
Every \method{} model spends the \emph{same total number of bits} as
NanoQuant. That is, at the 1 bpw budget,
0.973\,bpw for 0.6B and 0.986\,bpw for 1.7B and 4B, and at 0.8\,bpw, 0.773 / 0.783 / 0.793 / 0.794\,bpw for 0.6B / 1.7B / 4B / 8B.  At 0.55\,bpw the actual values are 0.527 / 0.535 / 0.544 / 0.545\,bpw, and the 8B curvature statistics use 128 sequences. Only the rank allocator
(\S\ref{sec:rank-alloc}) differs across sizes.

\textbf{Next Token Prediction.}
Table~\ref{tab:main} compares \method{} with NanoQuant at identical bit
budgets. PPL is lower at every model size, and the gap narrows as
the model grows. At 0.8\,bpw the PPL margin over NanoQuant widens at 1.7B and 4B (13 to 26\% and 3 to 11\%), narrows to 0.4\% at 8B, and at 0.6B and 1.7B \method{} at 0.8\,bpw matches or beats NanoQuant at 1.0\,bpw. Figure~\ref{fig:ppl-vs-size} shows both metrics across model sizes through 14B.  The right block reports the zero-shot mean. NanoQuant publishes zero-shot accuracies for Qwen3 only at 8B, and at 4B the rank allocator trades perplexity against zero-shot accuracy
(\S\ref{sec:exp-rank}).
\begin{table}[t]
\caption{WikiText-2 perplexity and zero-shot mean across a suite of reasoning tasks for $<$1\,bpw Qwen3-Base
models at the same total bits as NanoQuant \citep{nanoquant2025}, with both reported and reproduced values. 
The upper block uses a 1.0\,bpw budget, the middle block a 0.8\,bpw budget, and the lower block a 0.55\,bpw budget. Unreported results are marked with --.}
\label{tab:main}
\centering
\footnotesize
\setlength{\tabcolsep}{4pt}
\begin{tabular}{@{}rlccccc ccccc@{}}
\toprule
\multicolumn{2}{c}{} & \multicolumn{5}{c}{PPL $\downarrow$} & \multicolumn{5}{c}{0-shot mean $\uparrow$} \\
\cmidrule(lr){3-7}\cmidrule(lr){8-12}
bpw & Method & 0.6B & 1.7B & 4B & 8B & 14B & 0.6B & 1.7B & 4B & 8B & 14B \\
\midrule
\multirow{3}{*}{$\approx$1.0} & NanoQuant (paper)      & 27.56 & 19.21 & 14.29 & 12.47 & 10.92 & -- & -- & -- & 0.4894 & -- \\
& NanoQuant (repr.) & 29.21 & 18.76 & 14.86 & 12.38 & 11.49 & 0.387 & 0.427 & 0.455 & 0.474 & 0.498 \\
& \method{} & \textbf{22.96} & \textbf{16.72} & \textbf{13.80} & \textbf{11.82} & \textbf{10.91} & \textbf{0.427} & \textbf{0.448} & \textbf{0.464} & \textbf{0.497} & \textbf{0.506} \\
\midrule

\multirow{3}{*}{$\approx$0.8} & NanoQuant (paper)      & 33.79 & 25.31 & 19.33 & 14.83 & 12.88 & -- & -- & -- & -- & -- \\
&NanoQuant (repr.) & 42.22 & 22.37 & 17.99 & 15.63 & -- & 0.387 & 0.414 & \textbf{0.431} & 0.443 & -- \\
&\method{} & \textbf{27.51} & \textbf{18.84} & \textbf{17.27} & \textbf{14.77} & -- & \textbf{0.398} & \textbf{0.426} & 0.430 & \textbf{0.447} & -- \\

\midrule
\multirow{3}{*}{$\approx$0.55} & NanoQuant (paper)      & 52.94 & 33.74 & 32.86 & 20.04 & 17.06 & -- & -- & -- & -- & -- \\
& NanoQuant (repr.) & 62.65 & 32.40 & 39.49 & 22.49 & -- & 0.355 & 0.380 & \textbf{0.409} & \textbf{0.419} & -- \\
& \method{} & \textbf{38.80} & \textbf{25.63} & \textbf{24.24} & \textbf{18.21} & -- & \textbf{0.362} & \textbf{0.384} & 0.397 & 0.410 & -- \\
\bottomrule
\end{tabular}
\end{table}

\subsection{Rank Allocation at Bit Parity}
\label{sec:exp-rank}

Table~\ref{tab:rank-alloc} compares the allocators of \S\ref{sec:rank-alloc}
at the same total bits.  At 0.6B the measured sensitivity $\times$ depth prior
is best, 2.5 perplexity below uniform ranks.  At 4B the measured allocator
gives 13.80 against 13.55 from the hand table but restores the zero-shot mean
(0.463 vs.\ 0.450); both are single runs.  The depth prior alone thus does
not recover what the type weights in the hand table yields at that width, and the
two allocators trade perplexity against zero-shot accuracy.

\begin{table}[t]
\begin{minipage}[t]{0.50\linewidth}
\caption{Rank allocators at the same bpw.  WikiText-2 PPL with zero-shot mean.  ``Measured'' is
probed sensitivity $\times$ depth prior.}
\label{tab:rank-alloc}
\centering
\footnotesize
\setlength{\tabcolsep}{4pt}
\begin{tabular}{@{}lcc@{}}
\toprule
Qwen3-Base & Hand table & Measured \\
\midrule
4B   & \textbf{13.55} / 0.450 & 13.80 / \textbf{0.463} \\
\bottomrule
\end{tabular}

\vspace{0.5ex}

\caption{End-to-end wall-clock time for reproduced NanoQuant and \method{} on Qwen3-Base at $1$~bpw on 1$\times$ NVIDIA A100 80\,GB.}
\label{tab:wallclock}
\begin{tabular}{@{}lccc@{}}
\toprule
Size & NanoQuant & \method{} & Overhead \\
\midrule
0.6B & 1\,h\,12 & 1\,h\,59 & $65\%$ \\
1.7B & 1\,h\,18 & 2\,h\,15 & $73\%$ \\
4B   & 2\,h\,30 & 4\,h\,30 & $80\%$ \\
\bottomrule
\end{tabular}
\end{minipage}\hfill
\begin{minipage}[t]{0.46\linewidth}
\caption{Cumulative attribution at Qwen3-1.7B-Base, uniform ranks,
0.986\,bpw, WikiText-2 perplexity.}
\label{tab:ablation}
\centering
\footnotesize
\setlength{\tabcolsep}{4pt}
\begin{tabular}{@{}lc@{}}
\toprule
Configuration & PPL \\
\midrule
NanoQuant, published & 19.21 \\
NanoQuant, our reproduction & 18.76 \\
NKP statistics, untempered & 19.21 \\
$+$ tempering $p=\tfrac12$ & 17.46 \\
$+$ fresh input statistic and refresh & 17.28 \\
KL-Shampoo statistics, untempered & 17.04 \\
$+$ tempering $p=\tfrac12$ (\method{}) & \textbf{16.72} \\
\bottomrule
\end{tabular}
\end{minipage}
\end{table}

\subsection{Quantization Cost}
\label{sec:exp-cost}

\method{} uses the same deployment format as NanoQuant, so inference
kernels, memory footprint, and decoding throughput are unchanged.
The cost is incurred during quantization, as shown in Table~\ref{tab:wallclock}.
Appendix~\ref{app:complexity} gives the asymptotic accounting and
Appendix~\ref{app:impl} the timing breakdown and cost measures.

\subsection{Ablation Studies}
\label{sec:exp-ablation}

Table~\ref{tab:ablation} adds the modifications of \S\ref{sec:method}
cumulatively at 1.7B with uniform ranks.  Our reproduction of NanoQuant is
already 0.45 below the published number.  A dense NKP fit without tempering
gives nothing over the diagonal baseline. Tempering the same statistics comprises the
single largest step, and replacing the NKP fit by the KL-Shampoo fit
improves both the untempered and the tempered models.  Only the tempering
and rank-allocation steps are individually outside noise.
The fresh input statistic and periodic refresh of \S\ref{sec:locality} are a
consistent gain at every block boundary, though their aggregate effect in
Table~\ref{tab:ablation} is within single-run noise.  For the tempering
exponent, $p=1$ (none) and $p=1/4$ (Shampoo) were both worse than
$p=1/2$ at 1.7B.

\section{Conclusion}
\label{sec:conclusion}
We introduced \method{}, which replaces the diagonal reconstruction geometry in NanoQuant with a dense metric estimated by a KL-Shampoo-style Kronecker fit of the empirical Fisher Information matrix. We also identify useful processing techniques for robust application of the curvature estimated under uncertainty, including shrinkage, tempering, and periodic refreshes. The continuous ADMM updates of NanoQuant then become Sylvester equation solutions, which can be computed in
a cached eigenbasis, while other components of NanoQuant, including crucially
the packed format, remain unchanged, so inference can inherit the speed of the customized kernel.  At the
same total number of bits, \method{} lowers WikiText-2 PPL and increases zero-shot reasoning performance of $\lesssim 1$\,bpw
Qwen3-Base models from 0.6B to 14B.
Future work includes parallelizing the computation and using further approximations to the dense curvature statistics that yield fast matrix multiplication, both to reduce wall clock time. In addition, as the depth prior and calibration budget were tuned at 0.6B, tuning would likely help, and understanding scaling behavior is left as a promising direction for future investigation.

\section*{AI use statement}
In this work, we used generative AI tools for implementing methods.
We have not used generative AI tools for proposing or refining hypotheses, designing or providing feedback on research methodology or experiments, or interpreting results.
Generating synthetic data sets, helping develop theoretical models or conceptual frameworks, formulating mathematical claims, providing critical ingredients for proving mathematical claims, assisting in the writing of proofs, assisting with translation, cleaning and reformating datasets, and supporting qualitative and thematic data analysis are not applicable to this work.

Additionally, we used generative AI tools for creating scientific figures, suggesting experimental parameters, brainstorming, identifying some relevant literature, and some editing.
We have reviewed all AI-assisted work. LLM-generated code was verified via tests reviewed by humans prior to implementation. AI-generated figures were produced as TikZ that was manually revised by humans, and AI-generated writing was manually reviewed by humans.
We take responsibility for the final content of this work, including text, claims or artifacts produced with the aid of generative AI.

\section*{Ethics statement}
This work is intended to reduce the energy consumption required to run AI models, which has been a globally increasing concern due to scarcity of resources, limited production capacity, and environmental impact. On the surface, we might expect that reducing the energy cost of model inference would have a positive impact on these issues. However, this should be considered in the context of Jevon's paradox, where a decreasing energy cost for using an item results in increased aggregate usage due to more widespread access and adoption to the point that total energy consumption due to the item increases. This effect, while not guaranteed, remains a possibility here.

\section*{Reproducibility statement}
To aid with reproducibility, we provide source code with configuration files to run experiments found in the paper, which we intend to open source as part of the publication. The implementation details of the algorithms and relevant hyperparameters are contained within \S\ref{sec:method} and appendices. In addition, in the appendices we have included some discussion of directions that were unsuccessful in our explorations.

\bibliography{iclr2027_conference}
\bibliographystyle{iclr2027_conference}

\appendix

\section{Implementation Details}
\label{app:impl}
\paragraph{Hyperparameters.}
All runs follow the protocol used by NanoQuant wherever \S\ref{sec:method}
does not change it: 128 WikiText-2 calibration sequences of 2048 tokens
selected with seed 0 (512 for the curvature statistics at 8B and 14B), shrinkage $\gamma=0.2$, NanoQuant's robust token
clipping, $K=400$ LB-ADMM iterations with the linear $\rho_k$ schedule, two
FP16 scale vectors per layer, and
\textsc{TuneFP}/\textsc{TuneLatentSTE}/\textsc{TuneScalesKD} with learning
rates $10^{-4}/10^{-5}/10^{-6}$, batch sizes $4/1/1$, 8 epochs each (at 8B and 14B the first two stages use 2, 4, 6, and 8 epochs for the q and k, v and o, gate, and up and down layers) with
cosine decay, NanoQuant's weighted MSE block loss, and forward-KL scale-only
distillation.  \method{} adds three KL-Shampoo passes at calibration, the
tempering exponent $p=\tfrac12$, the refresh period $\Delta=7$ for the
28-block 0.6B and 1.7B models and $\Delta=9$ for the 36-block 4B model and 8B models, and $\Delta=10$ for the 40-block 14B model, the
fresh input statistic, and the rank allocator listed in Table~\ref{tab:main}.
The measured allocator probes each layer at
$\{0.5,1,1.5\}\times r^{\star}_\ell$ at 0.6B and 1.7B and at $\{0.6,1,1.4\}\times r^{\star}_\ell$ from 4B upward, with $K=50$.  Every $n\times n$ and
$m\times m$ eigendecomposition in the reported runs is computed in fp64.

\paragraph{Keeping the dense recipe within budget.}
The recipe costs more than NanoQuant per block: dense statistics per layer, one
eigendecomposition per statistic per ADMM call, the fresh input measurement,
the refresh passes, the rank probe, and late blocks whose allocated ranks
reach 2560 at 4B (\S\ref{sec:exp-cost}).  With allocated ranks at 4B, ADMM
accounts for 260\,s of a 473\,s block at block 7, and for about 40\,\% of
the block with uniform ranks.  Four measures keep this within budget.  (i) The fp32 statistics of all layers do not fit next to the model at
4B (about 36\,GB), so the calibration passes accumulate the Kronecker
statistics on the GPU over layer groups sized to a memory budget instead of
streaming to the CPU, reducing 4B calibration from about 5\,h to about
35\,min. The KL fit's per-pass inverses are Cholesky solves on the
accumulation device, reducing 1.7B calibration from 55 to 9\,min.  (ii) The
fresh input statistic of \S\ref{sec:locality} is measured once per shared-input group
and its eigendecomposition cached for the group's ADMM calls, saving three
of seven measurements and three large eigendecompositions per block.
(iii) The $n\times n$ eigendecompositions can run in fp32: the eigenvalues
are clamped and the result feeds a sign projection, and 1.7B perplexity was
unchanged (16.66 vs.\ 16.72 with fp64).  The reported 1.7B and 4B results
used fp64, so this is a cost measure only.  (iv) Calibration statistics, the
rank probe, every per-layer ADMM solution (content-addressed by weight bytes,
curvature tensors, and ADMM settings), per-block checkpoints, and the
pre-distillation model are cached, so an interrupted run resumes from its
last block in minutes.  Refresh costs are small, as demonstrated by how three refreshes cost
4\,min in total at 1.7B. The rank probe cost is given in
Appendix~\ref{app:rank}.

\section{Rank Allocation Details}
\label{app:rank}
This appendix completes \S\ref{sec:rank-alloc}.  Both allocators start
from the uniform ranks $r^{\mathrm{uni}}_\ell$ of Eq.~\ref{eq:uniform-rank},
keep their total $N_{\mathrm{bits}}$, and move ranks in steps of 32 with
ceiling $\min(m_\ell,n_\ell)$.

\paragraph{Hand table.}
The per-layer budget is $\beta_\ell=\beta\,\zeta\,\omega_\ell$ with the
depth prior $\omega_\ell$ of \S\ref{sec:rank-alloc} and type weights
$w=(\text{q}\ 0.85,\ \text{k}\ 0.90,\ \text{v}\ 1.10,\ \text{o}\ 1.00,\
\text{gate}\ 1.00,\ \text{up}\ 1.05,\ \text{down}\ 1.15)$,
set from the median relative block-loss jump each layer type caused when
binarized in the 4B logs, and $\zeta$ chosen so that
$\sum_\ell\beta_\ell m_\ell n_\ell=\beta\sum_\ell m_\ell n_\ell$.  Ranks are
obtained from $\beta_\ell$ by Eq.~\ref{eq:uniform-rank}, floored to multiples
of 32, and then moved in $\pm32$ steps, largest relative deviation from
$\beta_\ell$ first, until the total equals $N_{\mathrm{bits}}$.

\paragraph{Measured sensitivity times depth prior.}
After calibration, every layer is solved by a short run of the
dense-curvature LB-ADMM of \S\ref{sec:block-recon} ($K=50$) at the three
candidate ranks $\{0.5,1,1.5\}\times r^{\star}_\ell$ at 0.6B and 1.7B and $\{0.6,1,1.4\}\times r^{\star}_\ell$ from 4B upward, and the deployed matrix is scored by the Gauss--Newton weight error
\begin{equation}
J_\ell(r)
=\left\lVert\mathbf W_\ell-\widehat{\mathbf W}_\ell(r)\right\rVert
 _{\mathbf G_{\gamma,\ell},\mathbf A_{\gamma,\ell}}^{2},
\label{eq:probe-error}
\end{equation}
with the layer's own shrunken statistics before the trace normalization, so that
values are comparable across layers.  A power law
$J_\ell(r)=e^{a_\ell}r^{-\eta_\ell}$ is fitted through the three probes and
its level multiplied by the depth prior of \S\ref{sec:rank-alloc} with all
$w_{t}=1$.  Starting from rank 32 everywhere, 32-rank steps are handed to the
layer with the largest predicted loss decrease per added bit,
$\big(J_\ell(r)-J_\ell(r+32)\big)/\big(32(m_\ell+n_\ell)\big)$, until
$N_{\mathrm{bits}}$ is reached.  The measured curves rank the seven layer
types within a block well (Spearman $0.8$--$0.96$ against logged block-loss
jumps) but under-value late blocks, which is what the prior supplies.  The
probe shares each statistic's eigendecomposition across its three candidate
ranks and is cached, taking 707\,s at 0.6B.

\section{Algorithms}
\label{app:algorithms}
This appendix lists the procedures of \S\ref{sec:block-recon} and the
unchanged NanoQuant components they call (\S\ref{sec:bg-nanoquant}), written
in the transformed coordinates.

\paragraph{Diagonal maps and transformed target (Step~2-1).}
With the maximum applied elementwise, the diagonal preconditioners are
\begin{align}
\mathbf d_{\mathrm{in}}
  &=\sqrt{\max\{\operatorname{diag}(\mathbf A_\gamma),
                         \epsilon\mathbf 1_n\}},
&
\widetilde{\mathbf D}_{\mathrm{in}}
  &=\Diag(\mathbf d_{\mathrm{in}}),
\nonumber\\
\mathbf d_{\mathrm{out}}
  &=\sqrt{\max\{\operatorname{diag}(\mathbf G_\gamma),
                         \epsilon\mathbf 1_m\}},
&
\widetilde{\mathbf D}_{\mathrm{out}}
  &=\Diag(\mathbf d_{\mathrm{out}}),
\label{eq:diag-coordinates}
\end{align}
and the transformed target, transformed error, and cached metric target are
\begin{equation}
\widetilde{\mathbf W}
=\widetilde{\mathbf D}_{\mathrm{out}}\mathbf W
 \widetilde{\mathbf D}_{\mathrm{in}},
\qquad
\widetilde{\mathbf E}
=\widetilde{\mathbf D}_{\mathrm{out}}\mathbf E
 \widetilde{\mathbf D}_{\mathrm{in}},
\qquad
\mathbf M
\mathrel{:=}
\overline{\mathbf G}\,\widetilde{\mathbf W}\,\overline{\mathbf A},
\label{eq:transformed-target}
\end{equation}
where $\mathbf M$ uses the tempered transported statistics of
Eq.~\ref{eq:tempering} and is cached together with their eigensystems.
Algorithm~\ref{alg:prepare-metric} lists the step.

\begin{algorithm}[H]
\caption{\textsc{PrepareMetricTarget} for one layer (Step~2-1).}
\label{alg:prepare-metric}
\begin{algorithmic}[1]
\REQUIRE $\mathbf W$, shrunken statistics $(\mathbf A_\gamma,\mathbf G_\gamma)$,
  with $\mathbf A_\gamma$ from the fresh measurement of
  \S\ref{sec:precondition}, floor $\epsilon$, tempering exponent $p$
\STATE Form $(\widetilde{\mathbf D}_{\mathrm{in}},
  \widetilde{\mathbf D}_{\mathrm{out}})$ by
  Eq.~\ref{eq:diag-coordinates}
\STATE $\widetilde{\mathbf W}\gets
  \widetilde{\mathbf D}_{\mathrm{out}}\mathbf W
  \widetilde{\mathbf D}_{\mathrm{in}}$
\STATE Transport to $(\overline{\mathbf A},\overline{\mathbf G})$ by
  Eq.~\ref{eq:metric-transport}, do not refit
\STATE Eigendecompose $\overline{\mathbf A},\overline{\mathbf G}$ and temper
  both by Eq.~\ref{eq:tempering}
\STATE $\mathbf M\gets
  \overline{\mathbf G}\widetilde{\mathbf W}\overline{\mathbf A}$,
  cache the tempered eigensystems
\RETURN $\widetilde{\mathbf W}$, both diagonal maps, and the tempered
  metric state
\end{algorithmic}
\end{algorithm}

\paragraph{Continuous block updates (Step~2-2).}
Set $\alpha_k=\rho_k+\lambda$.  Holding $\mathbf V^k$ fixed, define
\begin{equation}
\mathbf R_V^k
=(\mathbf V^k)^\top\overline{\mathbf A}\mathbf V^k,
\qquad
\mathbf B_U^k
=\mathbf M\mathbf V^k+\rho_k\mathbf C_U^k,
\label{eq:u-system-parts}
\end{equation}
so that Eq.~\ref{eq:u-sylvester} reads
$\overline{\mathbf G}\,\mathbf U^{k+1}\mathbf R_V^k
+\alpha_k\mathbf U^{k+1}=\mathbf B_U^k$.  After this update, define
\begin{equation}
\mathbf R_U^{k+1}
=(\mathbf U^{k+1})^\top
  \overline{\mathbf G}\mathbf U^{k+1},
\qquad
\mathbf B_V^k
=\mathbf M^\top\mathbf U^{k+1}+\rho_k\mathbf C_V^k,
\label{eq:v-system-parts}
\end{equation}
which yields
\begin{equation}
\overline{\mathbf A}\,\mathbf V^{k+1}\mathbf R_U^{k+1}
+\alpha_k\mathbf V^{k+1}
=\mathbf B_V^k.
\label{eq:v-sylvester}
\end{equation}
Both systems are solved by Eq.~\ref{eq:sylvester-solve}, and these two solves are denoted by
\textsc{KronUSolve} and \textsc{KronVSolve} in
Algorithm~\ref{alg:kron-admm}.

\paragraph{Euclidean proxy projection.}
For a proxy $\mathbf P\in\mathbb R^{d\times r}$, SVID computes
\begin{equation}
\mathbf S=\sign(\mathbf P),\qquad
(\mathbf a,\sigma,\mathbf b)
=\textsc{TopSingularTriplet}(|\mathbf P|),
\label{eq:svid-triplet}
\end{equation}
with nonnegative unit vectors $\mathbf a,\mathbf b$ and zeros in
$\mathbf P$ mapped to $+1$ by $\sign$.  Defining
$\mathbf p=\sqrt{\sigma}\mathbf a$ and
$\mathbf q=\sqrt{\sigma}\mathbf b$, the projected proxy is
\begin{equation}
\textsc{SVID}(\mathbf P)
=(\sigma\mathbf a\mathbf b^\top)\odot\mathbf S
=\Diag(\mathbf p)\mathbf S\Diag(\mathbf q).
\label{eq:svid-projection}
\end{equation}
The triplet is obtained by the same power iteration as NanoQuant.  Since the
consensus penalties in Eq.~\ref{eq:kron-admm-objective} are Euclidean,
neither Eq.~\ref{eq:svid-projection} nor its inner iteration count depends on
$\overline{\mathbf A}$ or $\overline{\mathbf G}$.

\paragraph{Scale extraction (Step~2-3).}
The final proxies $\mathbf P_U,\mathbf P_V$ live in the transformed
coordinates of Eq.~\ref{eq:transformed-target}.  NanoQuant's magnitude
balancing (its Eq.~7) and scale extraction (its Eq.~8) are applied to them
verbatim, yielding sign matrices $\mathbf S_U,\mathbf S_V$ and transformed
scale vectors $\widetilde{\mathbf s}_1,\widetilde{\mathbf s}_2$.  The return
to the original weight coordinates is
\begin{equation}
\mathbf s_1=\mathbf d_{\mathrm{out}}^{-1}\odot\widetilde{\mathbf s}_1,
\qquad
\mathbf s_2=\mathbf d_{\mathrm{in}}^{-1}\odot\widetilde{\mathbf s}_2,
\label{eq:two-scales}
\end{equation}
which is exactly NanoQuant's un-preconditioning step with the diagonal maps
of Eq.~\ref{eq:diag-coordinates}.  \textsc{Finalize} in
Algorithm~\ref{alg:kron-admm} denotes these operations.

\begin{algorithm}[H]
\caption{\textsc{Kron-LB-ADMM} for one transformed layer (Step~2-2).}
\label{alg:kron-admm}
\begin{algorithmic}[1]
\REQUIRE Prepared state from Algorithm~\ref{alg:prepare-metric}, rank $r$,
  iterations $K$, penalty schedule $\rho_k$, ridge $\lambda$
\STATE Initialize $\mathbf U^0,\mathbf V^0$ as in NanoQuant
\STATE $\mathbf Z_U^0\gets\textsc{SVID}(\mathbf U^0)$,\quad
  $\mathbf Z_V^0\gets\textsc{SVID}(\mathbf V^0)$
\STATE Initialize scaled duals
  $\boldsymbol\Lambda_U^0,\boldsymbol\Lambda_V^0$
\FOR{$k=0,\ldots,K-1$}
  \STATE $\mathbf C_U^k\gets
    \mathbf Z_U^k-\boldsymbol\Lambda_U^k$,\quad
    $\mathbf C_V^k\gets
    \mathbf Z_V^k-\boldsymbol\Lambda_V^k$
  \STATE $\mathbf U^{k+1}\gets
    \textsc{KronUSolve}(\mathbf V^k,\mathbf C_U^k;
      \overline{\mathbf G},\overline{\mathbf A},\mathbf M,
      \rho_k,\lambda)$
  \STATE $\mathbf V^{k+1}\gets
    \textsc{KronVSolve}(\mathbf U^{k+1},\mathbf C_V^k;
      \overline{\mathbf G},\overline{\mathbf A},\mathbf M,
      \rho_k,\lambda)$
  \STATE $\mathbf Z_U^{k+1}\gets
    \textsc{SVID}(\mathbf U^{k+1}+\boldsymbol\Lambda_U^k)$
  \STATE $\mathbf Z_V^{k+1}\gets
    \textsc{SVID}(\mathbf V^{k+1}+\boldsymbol\Lambda_V^k)$
  \STATE $\boldsymbol\Lambda_U^{k+1}\gets
    \boldsymbol\Lambda_U^k+\mathbf U^{k+1}-\mathbf Z_U^{k+1}$
  \STATE $\boldsymbol\Lambda_V^{k+1}\gets
    \boldsymbol\Lambda_V^k+\mathbf V^{k+1}-\mathbf Z_V^{k+1}$,
    update Euclidean residuals
\ENDFOR
\STATE $\mathbf P_U\gets
  \mathbf U^K+\boldsymbol\Lambda_U^K$,\quad
  $\mathbf P_V\gets
  \mathbf V^K+\boldsymbol\Lambda_V^K$
\RETURN $\textsc{Finalize}(\mathbf P_U,\mathbf P_V,
  \widetilde{\mathbf D}_{\mathrm{out}},
  \widetilde{\mathbf D}_{\mathrm{in}})$
\end{algorithmic}
\end{algorithm}

\paragraph{End-to-end pipeline.}
Algorithm~\ref{alg:end-to-end} shows the complete pipeline.  Calibration
statistics are first collected from the full-precision teacher, so every
initial KL-Shampoo fit is defined in the original layer coordinates, and
ranks are allocated once.  After \textsc{TuneFP} and the fresh measurement,
$\widetilde{\mathbf W}$, the transported and tempered statistics, and
$\mathbf M$ are recomputed for the current target.  Latent tuning and final
knowledge distillation act on
$\mathcal U,\mathcal V,\mathbf s_1,\mathbf s_2$ exactly as in NanoQuant.

\begin{algorithm}[H]
\caption{Dense-curvature \method{}.}
\label{alg:end-to-end}
\begin{algorithmic}[1]
\REQUIRE Full-precision teacher $\mathcal M$, calibration set
  $\mathcal X_{\mathrm{cal}}$, bit budget $\beta$ and allocator
  (\S\ref{sec:rank-alloc}), $\gamma,\epsilon,p,\Delta$,
  ADMM parameters $K,\rho,\lambda$, tuning budgets
  $T_{\mathrm{pre}},T_{\mathrm{post}},T_{\mathrm{glob}}$
\STATE \textit{// Phase 1: global calibration and rank allocation}
\STATE Collect $(\mathbf x_t,\boldsymbol\delta_t)$ for every linear layer
  on $\mathcal X_{\mathrm{cal}}$ through $\mathcal M$
\FOR{each linear layer $\ell$}
  \STATE $(\mathbf A,\mathbf G)^{(\ell)}\gets
    \textsc{KL-Shampoo}$ by Eq.~\ref{eq:kl-fixed-point}, three passes,
    shrink by Eq.~\ref{eq:kron-shrink}
\ENDFOR
\STATE Allocate ranks $\{r_\ell\}$ at bit parity
  (\S\ref{sec:rank-alloc}), probing with $K=50$ if the measured allocator is used
\STATE \textit{// Phase 2: block reconstruction}
\STATE $\widehat{\mathcal M}\gets\mathcal M$
\FOR{blocks $b=1,\ldots,B$}
  \IF{$b>1$ and $(b-1)\bmod\Delta=0$}
    \STATE Refresh $(\mathbf A,\mathbf G)^{(\ell)}$ of all unfactorized
      layers with one warm-started pass of Eq.~\ref{eq:kl-fixed-point}
      through $\widehat{\mathcal M}$, shrink
  \ENDIF
  \STATE $\mathbf X_b\gets
    \widehat{\mathcal M}_{<b}(\mathcal X_{\mathrm{cal}})$,\quad
    $\mathbf Y_b\gets\mathcal B_b^{\mathrm{FP}}(\mathbf X_b)$
  \STATE $\textsc{TuneFP}(b,\mathbf X_b,\mathbf Y_b;T_{\mathrm{pre}})$
    \hfill\textit{// Step 1}
  \FOR{shared-input groups $g\in b$ (q\,k\,v, o, gate\,up, down)}
    \STATE Measure $\mathbf A_{\mathrm{fresh}}^{(g)}$
      (\S\ref{sec:precondition}) on the current inputs of $g$, shrink
      by Eq.~\ref{eq:kron-shrink}
    \FOR{linear layers $\ell\in g$ with weight $\mathbf W^{(\ell)}$}
      \STATE $\textsc{PrepareMetricTarget}$ (Algorithm~\ref{alg:prepare-metric})
        with $\mathbf A_\gamma\gets\mathbf A_{\mathrm{fresh},\gamma}^{(g)}$
        and the current $\mathbf G_\gamma^{(\ell)}$
        \hfill\textit{// Step 2-1}
      \STATE $(\mathcal U,\mathcal V,\mathbf s_1,\mathbf s_2)^{(\ell)}
        \gets\textsc{Kron-LB-ADMM}$ at rank $r_\ell$ by
        Algorithm~\ref{alg:kron-admm}
        \hfill\textit{// Steps 2-2, 2-3}
    \ENDFOR
  \ENDFOR
  \STATE $\textsc{TuneLatentSTE}
    (b,\mathbf X_b,\mathbf Y_b;T_{\mathrm{post}})$ over
    $\mathcal U,\mathcal V,\mathbf s_1,\mathbf s_2$
    \hfill\textit{// Step 3}
  \FOR{linear layers $\ell\in b$}
    \STATE $\mathbf U_{\pm1}^{(\ell)}
      \gets\sign(\mathcal U^{(\ell)})$,\quad
      $\mathbf V_{\pm1}^{(\ell)}
      \gets\sign(\mathcal V^{(\ell)})$
    \STATE $\textsc{PackBinary}
      (\mathbf U_{\pm1}^{(\ell)},\mathbf V_{\pm1}^{(\ell)})$
  \ENDFOR
\ENDFOR
\STATE \textit{// Phase 3: model reconstruction}
\STATE $\textsc{TuneScalesKD}
  (\widehat{\mathcal M},\mathcal M,\mathcal X_{\mathrm{cal}};
   T_{\mathrm{glob}})$ over $\mathbf s_1,\mathbf s_2$
\RETURN $\widehat{\mathcal M}$
\end{algorithmic}
\end{algorithm}

\section{Complexity Analysis}
\label{app:complexity}
Table~\ref{tab:complexity} separates calibration, preparation, and one outer
ADMM sweep for a layer with $N$ calibration tokens.  The dense setup includes
forming $\mathbf M$ and eigendecomposing the two transported statistics.
\begin{table}[H]
\caption{Per-layer costs for an $m\times n$ weight, factor rank $r$, and $N$
calibration tokens.}
\label{tab:complexity}
\begin{center}
\begin{tabular}{p{0.20\linewidth}p{0.30\linewidth}p{0.40\linewidth}}
\toprule
Component & NanoQuant diagonal geometry & \method{} dense geometry \\
\midrule
Curvature state
& $\Theta(m+n)$
& $\Theta(m^2+n^2)$ statistics and eigensystems \\
Calibration fit
& $\Theta(N(m+n))$
& $\Theta(N(m^2+n^2)+m^3+n^3)$ per pass, three passes \\
Fresh input statistic
& --
& $\Theta(Nn^2)$ per shared-input group \\
Refresh
& --
& one fit pass over unfactorized layers every $\Delta$ blocks \\
Rank probe
& --
& three $K{=}50$ ADMM solves per layer, once \\
Setup
& $\Theta(mn)$
& $\Theta(m^3+n^3+m^2n+mn^2)$ \\
One outer sweep
& $\Theta(mnr+(m+n)r^2+r^3)$
& $\Theta(mnr+(m^2+n^2)r+(m+n)r^2+r^3)$ \\
SVID inner work
& $\Theta(K_{\mathrm{in}}(m+n)r)$
& unchanged \\
\bottomrule
\end{tabular}
\end{center}
\end{table}

Each layer performs two continuous solves per outer iteration and two SVID
projections at initialization plus two per iteration.  For representation
$x\in\{\mathrm{diagonal},\mathrm{dense}\}$, its reconstruction runtime can
be decomposed as
\begin{equation}
T_{x,\ell}
=S_{x,\ell}
+K_{\mathrm{out},\ell}C_{x,\ell}^{\mathrm{outer}}
+(K_{\mathrm{out},\ell}+1)C_\ell^{\mathrm{SVID}}
+N_{\mathrm{diag},\ell}D_{x,\ell},
\label{eq:runtime-accounting}
\end{equation}
where $S$ is setup, $C^{\mathrm{outer}}$ is one pair of continuous solves,
$C^{\mathrm{SVID}}$ is one pair of proxy projections, and $D$ covers
diagnostic evaluations.  This accounting makes the one-time dense setup
explicit rather than hiding it in the iteration cost.  In practice, the
$(m^2+n^2)r$ term does not dominate: at 4B, where allocated ranks reach
$r\approx1500$--$2500$ and each layer performs 800 $r\times r$
eigendecompositions over $K=400$ iterations, removing every $n^2r$ product
changes per-layer ADMM time by only 5--8\,\%, so the per-iteration
$r\times r$ eigendecomposition dominates instead (Appendix~\ref{app:tried}).

\section{Design Alternatives Not Adopted}
\label{app:tried}
The following variants were evaluated and are not part of \method{}.

\textbf{Variants within noise.}
A dense Mahalanobis \emph{block} loss for \textsc{TuneFP} and
\textsc{TuneLatentSTE} (NanoQuant's weighted MSE is kept), a spike-plus-flat
projection of the curvature statistics, distilling the latent binaries in the KD stage,
feature distillation, trainable norm weights, and best-epoch selection in the
KD stage, a logit-level objective for the last blocks, lifting the rank
ceiling, and spending the rounding remainder to reach exactly 1.0\,bpw were
all within single-run noise.  An explicit per-rank middle scale vector
between the two sign matrices (\S\ref{sec:scales}), carried from ADMM through
tuning and KD, was 3.5--5 perplexity \emph{worse} at 0.6B: a mean-one
rebalanced export of the same vector was within noise, and a sweep of its
learning rate was flat to negative.  On the 0.6B four-block screen used for
tuning-budget changes (block-3 perplexity 14.15 for the control, single-run
noise about 0.2), a plateau stop that ends an epoch loop once the block loss
improves by less than $10^{-2}$ was within noise (14.23), while running one
\textsc{TuneFP} round per shared-input group (four rounds per block instead
of seven) cost $+0.28$, outside noise, so every round matters, including
those between layers that read the same activation.

\textbf{Speed variants.}
An inexact Sylvester solve (a stale eigenbasis as preconditioner with
Rayleigh, QR, and conjugate-gradient rungs) saved about 10\,\% of ADMM time
at 1.7B and nothing at 4B, and early stopping of ADMM never fires: at
iteration 400 every layer still flips between 30 and 20{,}000 signs per
iteration under the linear $\rho_k$ schedule.  A low-rank-plus-identity
Sylvester step for spectrally projected statistics, which keeps the
$k_{\mathrm s}$ largest and $k_{\mathrm d}$ smallest eigenvalues of each
transported statistic exactly and replaces the rest by their geometric mean so
that each update costs $O(nkr+nr^2)$ instead of $O(n^2r)$ with
$k=k_{\mathrm s}+k_{\mathrm d}$, saved only 5--8\,\% of 4B ADMM time, because
the $r\times r$ eigendecomposition and not the $n^2r$ products dominates at
those widths (Appendix~\ref{app:complexity}): the projection itself cost
$+0.10$ to $+0.34$ perplexity at block 3 with $k=64$ per side and was within
noise at $k=256$.  It is not used.

\end{document}